\documentclass[sigconf, nonacm]{acmart}
\AtBeginDocument{%
  }

\setcopyright{acmlicensed}
\copyrightyear{2025}
\acmYear{2025}
\acmDOI{XXXXXXX.XXXXXXX}
\acmConference[CIKM '25]{Proceedings of the 34th ACM International Conference on Information and Knowledge Management}{November 10--14, 2025}{Seoul, Korea}  
\acmISBN{978-1-4503-XXXX-X/2018/06}

\usepackage{graphicx}
\usepackage{multirow}
\begin{document}

\title{Generating Narrated Lecture Videos from Slides with Synchronized Highlights}

\author{Alexander Holmberg}
\affiliation{%
  \institution{KTH Royal Institute of Technology}
  \city{Stockholm}
  \country{Sweden}}
\email{alholmbe@kth.se}

\renewcommand{\shortauthors}{Holmberg}

\begin{abstract}
Turning static slides into engaging video lectures takes considerable time and effort, requiring presenters to record explanations and visually guide their audience through the material. We introduce an end-to-end system designed to automate this process entirely. Given a slide deck, this system synthesizes a video lecture featuring AI-generated narration synchronized precisely with dynamic visual highlights. These highlights automatically draw attention to the specific concept being discussed, much like an effective presenter would. The core technical contribution is a novel highlight alignment module. This module accurately maps spoken phrases to locations on a given slide using diverse strategies (e.g., Levenshtein distance, LLM-based semantic analysis) at selectable granularities (line or word level) and utilizes timestamp-providing Text-to-Speech (TTS) for timing synchronization. We demonstrate the system's effectiveness through a technical evaluation using a manually annotated slide dataset with 1000 samples, finding that LLM-based alignment achieves high location accuracy (F1 > 92\%), significantly outperforming simpler methods, especially on complex, math-heavy content. Furthermore, the calculated generation cost averages under \$1 per hour of video, offering potential savings of two orders of magnitude compared to conservative estimates of manual production costs. This combination of high accuracy and extremely low cost positions this approach as a practical and scalable tool for transforming static slides into effective, visually-guided video lectures.
\end{abstract}

  
\begin{teaserfigure}
  \includegraphics[width=\textwidth]{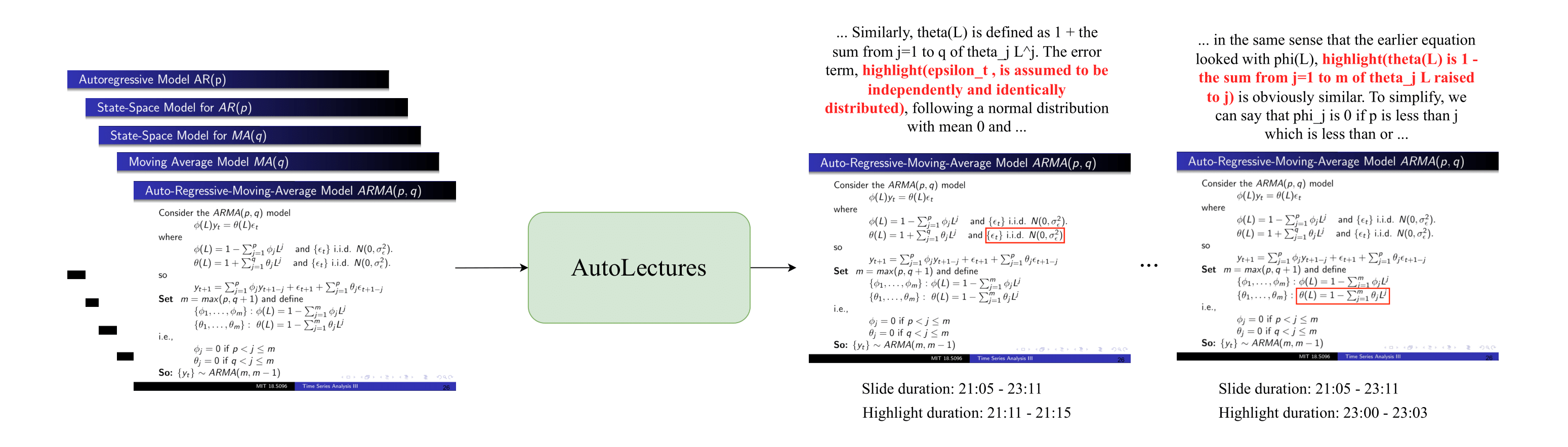}
  \caption*{Overview of AutoLectures. Input slides are processed to generate a video lecture where different spoken concepts in the narration trigger accurately timed visual highlights on corresponding elements of the slide. Slide example is sourced from MIT 18.S096 Topics in Mathematics with Applications in Finance \cite{mit_18s096_ocw}.}
  \label{fig:teaser}
  \Description{A wide figure illustrating the AutoLectures workflow. On the left, an example input slide shows calculating cross-entropy loss. An arrow points to a central box labeled 'AutoLectures'. An arrow points from this box to the right side, which shows three versions of the input slide overlaid with different narration excerpts and corresponding red highlight boxes. One highlights 'cross-entropy loss' over the formula, another highlights 's = Wx + b' over that equation, and the third highlights 'normalized score' over the relevant text/numbers. Timing information is shown below each highlighted example.}
\end{teaserfigure}


\maketitle

\section{Introduction}
\label{sec:introduction}

Video lectures are an essential part of modern education, offering enhanced engagement, accessibility, and flexibility for learners across various settings, from university courses and MOOCs to corporate training. While presentation slides are readily available precursors to these lectures, they remain static artifacts, lacking the dynamic narration and crucial visual guidance that facilitate comprehension. Creating high-quality video lectures manually, however, is a resource-intensive endeavor. It demands valuable educator time and effort for recording, editing, and updates, diverting focus from primary activities like research and direct student interaction.

Furthermore, static slides and unguided video lectures often fail to leverage a key principle of effective multimedia learning: the \textbf{Signalling Principle} (also known as the Cueing Principle)~\cite{signallingprinciple}. This principle highlights that learners benefit significantly when their attention is actively guided towards essential information precisely when it is relevant. Effective human presenters achieve this naturally using gestures or annotations. Automatically generating video presentations from slides that replicate not only the narration but also these timed visual cues presents a technical challenge, and overcoming this is key to creating automated lectures that are not just narrated, but also pedagogically effective.

To address this gap, we present AutoLectures, an end-to-end system designed for the automated synthesis of narrated video lectures directly from PDF slide decks. AutoLectures aims to produce videos that are not only narrated but also incorporate dynamically synchronized visual highlights. This feature seeks to mimic the attention-guiding visual cues commonly employed by human lecturers, thereby enhancing the quality and effectiveness of the automatically generated content without manual intervention. While human presenters apply visual cues intuitively, automatically determining what textual elements correspond to the spoken narration, where these elements are precisely located on slides with varying layouts (including text, figures, and mathematical notation), and when to display highlights in perfect synchrony with synthesized speech poses algorithmic challenges. AutoLectures tackles these challenges using a multi-stage processing pipeline that integrates different components, including Large Language Models for script generation, Optical Character Recognition for layout analysis, and timestamp-providing Text-to-Speech models for synchronized audio, along with a highlight alignment module. 

The cornerstone of our system is the configurable highlight alignment module. It addresses the `where` (location) and `when` (timing) challenges for synchronized highlights. Location is handled by offering choices in matching granularity (`line` or `word` level OCR elements) and method (e.g., `simple`, `fuzzy`, or `LLM`-based semantic matching). Timing relies on word-level timestamps from the TTS service (Section \ref{sec:timing_lookup}). This configurability allows the system to adapt to different content types (e.g., text-heavy vs. math-heavy slides) and user preferences for accuracy versus cost. In this paper, we present the following core contributions:
\begin{enumerate}
    \item The design and implementation of AutoLectures: An end-to-end system automating the synthesis of narrated video lectures with synchronized visual highlights directly from PDF slide decks.
    \item A novel configurable highlight alignment module integrating diverse location matching strategies (including LLM-based) and granularities with precise, TTS-derived timing for effective highlight generation.
    \item A comprehensive technical evaluation assessing highlight location accuracy (Section \ref{sec:eval_word_location_accuracy}), system performance, and cost efficiency (Section \ref{sec:cost_analysis}) across diverse slide types. This evaluation is grounded in \textbf{AutoLectures-1K}, a new dataset we created containing 1000 manually annotated word-level highlight instances with ground-truth visual polygons, which we also release.
\end{enumerate}

\section{Related Work}
\label{sec:related_work}

\subsection{Automated Slide-to-Video Generation}

Automating aspects of presentations has been explored from different angles. One line of research focuses on generating presentation slides directly from source documents, often academic papers or general text. \textit{PASS} \cite{aggarwal2025passpresentationautomationslide} generates both slides and corresponding AI narration from general documents. Another line of research focuses on enhancing existing slides by automatically adding narration or avatars. \textit{AutoLV}~\cite{autolv}, for example, synthesizes voice-overs and talking heads for pre-annotated slide decks.

A common limitation unites these different approaches: the lack of automatically generated, dynamic visual guidance synchronized with the narration. While systems like PASS produce narrated slides and AutoLV adds audio to existing ones, they do not incorporate mechanisms to actively guide the viewer's attention to specific textual content precisely when it is being discussed. Consequently, learners using videos produced by such systems are often still required to locate relevant information themselves on potentially complex slides, diminishing the effectiveness compared to a presentation with visual cues.

\subsection{Dynamic Visual Cueing in Multimedia Learning}

Beyond the basic generation of slides and narration, the pedagogical effectiveness of multimedia presentations is significantly influenced by how viewer attention is managed. Educational psychology, particularly within the framework of multimedia learning theory, highlights the value of directing a learner’s visual attention toward essential information precisely when it is being discussed. The Signalling Principle (or Cueing Principle) ~\cite{signallingprinciple} encapsulates this finding, stating that learners benefit significantly when cues are added to highlight key material and its organization. Such cues, which can include visual highlighting, arrows, color-coding, or vocal emphasis, serve to reduce extraneous cognitive load by minimizing the learner's need to search for relevant information. By guiding attention effectively, signals allow learners to better focus cognitive resources on understanding and integrating the core content, leading to improved retention and transfer. For instance, recent experimental work using pedagogical agents demonstrated that incorporating specific, synchronized visual cues, particularly pointing gestures aligned with narration, significantly improves learning outcomes and directs learners' visual fixations compared to conditions lacking such guidance~\cite{Li_Wang_Mayer_2023}. The empirical support for this principle underscores the importance of incorporating synchronized, attention-guiding mechanisms into the design of effective instructional videos, a capability often overlooked in automated presentation generation tools.

\section{The AutoLectures System}
\label{sec:system_overview}

AutoLectures transforms a given PDF slide deck into a dynamic video presentation featuring synthesized narration synchronized with timed textual highlights. The system operates via a multi-stage processing pipeline, illustrated in Figure \ref{fig:system_architecture}. This pipeline processes each slide, leveraging Large Language Models (LLMs), Optical Character Recognition (OCR), and Text-to-Speech (TTS) modules to generate the necessary elements for the final video assembly. The core stages of the pipeline are as follows:

\begin{figure}[t] 
  \centering
\includegraphics[width=\linewidth]{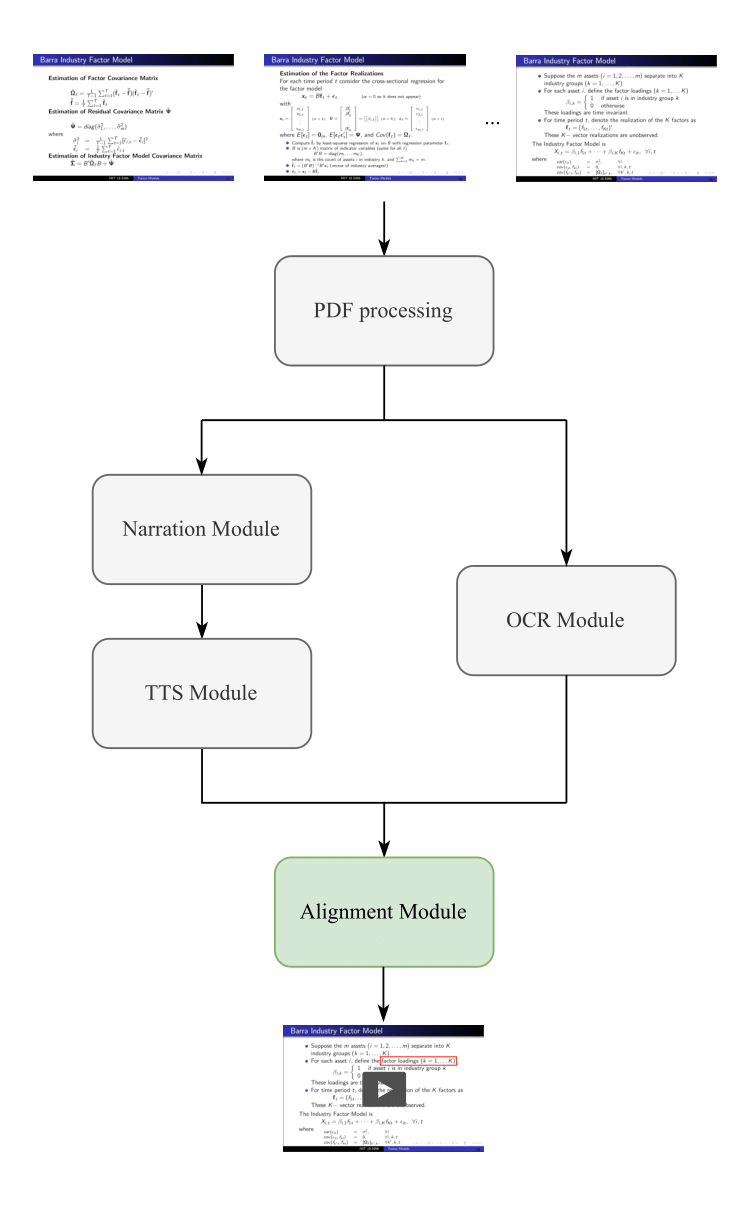}
  \caption{The AutoLectures system architecture. Starting with a PDF input, the pipeline performs PDF processing, then branches into Narration and OCR modules. Narration output feeds a TTS module. Outputs from TTS and OCR feed the core Alignment Module, which produces the final video output.} 
  \label{fig:system_architecture}
  \Description{A flowchart diagram illustrating the AutoLectures pipeline. Three slide thumbnails feed into a 'PDF processing' box. This box outputs to both a 'Narration Module' box on the left and an 'OCR Module' box on the right. The 'Narration Module' outputs to a 'TTS Module' box below it. Both the 'TTS Module' and the 'OCR Module' output arrows pointing to a central 'Alignment Module' box below them, highlighted in green. The 'Alignment Module' outputs an arrow pointing to a video player thumbnail representing the final video output.}
\end{figure}

\subsection{Narration Module}
\label{sec:narration_generation}
For each slide image extracted from the input PDF, a Large Language Model (LLM) generates a narration script suitable for speech synthesis. We instruct the LLM to explain the slide's content comprehensively, while ensuring narrative continuity if processing multiple slides. This involves not just relaying text but also describing significant visual elements (diagrams, graphs) and explaining the meaning or purpose behind mathematical notation rather than merely reading symbols aloud. As part of this generation, the LLM identifies key terms, definitions, or specific formulas discussed in its narration and embeds special `highlight()` markers around the corresponding text exactly as it appears visually on the slide (e.g., "...uses highlight(gradient descent) to find..."). These markers designate the content for visual emphasis in the final video. The resulting transcript, containing the narration and highlight markers, drives both the audio synthesis and highlight alignment stages.

\subsection{OCR Module}
\label{sec:ocr}
Each slide image is processed by an OCR module to extract detailed layout information. This includes the precise coordinates for individual text lines and words, which are essential for accurate highlight placement during the alignment phase.

\subsection{TTS Module}
\label{sec:tts}
The narration script (with highlight() markers removed) is next fed into a TTS module. AutoLectures requires a TTS model capable of outputting not only the audio waveform but also reliable, word-level timestamps indicating the start and end time of each spoken word. This direct timing information is key for the highlight synchronization process.

\subsection{Alignment Module}
\label{sec:highlight_alignment_overview}
This central module integrates the outputs from the previous stages: the `highlight()` markers from the transcript, the geometric data from the OCR module, and the word-level timestamps from the TTS module. Its core function is to map each highlighted phrase to its corresponding visual location(s) on the slide and determine the precise time interval (start/end milliseconds) for its display, synchronized with the audio narration. The output is a set of render events for the final video stage. The detailed mechanisms and configurability of this module are presented in Section \ref{sec:highlight_alignment}.

\subsection{Video Synthesis}
\label{sec:video_synthesis}
In the final stage, the video lecture is rendered. For each slide, the original image, the synthesized audio segment, and the calculated highlight render events are combined. Using video processing tools, the slide image is displayed, the narration is overlaid, and the highlight bounding boxes are drawn dynamically, appearing and disappearing according to the precise timing specified by the alignment module. The per-slide segments are then concatenated to create the final video output.

\section{Configurable Highlight Alignment}
\label{sec:highlight_alignment}
The Highlight Alignment module is the technical core responsible for translating the abstract `highlight()` markers embedded in the narration script (Section \ref{sec:narration_generation}) into concrete visual events synchronized with the audio. This involves tackling two main challenges: determining the precise spatial location (\textit{where}) on the slide image the highlighted text resides, and calculating the exact temporal interval (\textit{when}) the highlight should be visible to match the corresponding speech segment produced by the TTS module (Section \ref{sec:tts}). Accurately resolving spatial ambiguity (e.g., repeated terms on a slide) and temporal synchronization is critical for producing effective visual guidance. To address variations in slide content (e.g., standard text vs. mathematical notation) and accommodate different priorities regarding accuracy, speed, and cost, this module is designed with key points of configurability, detailed below. The module takes as input the `highlight(phrase)` tokens, the geometric data from OCR (Section \ref{sec:ocr}), and the word-level timestamps from the TTS module (Section \ref{sec:tts}), and outputs a list of render events, each specifying highlight polygons and a precise time interval.

\subsection{Location Matching}
\label{sec:location_matching}

The primary goal of location matching is, for a given slide $s$, to identify the specific region(s) on the slide image that visually correspond to a phrase $p$ marked for highlighting in the transcript. The input consists of the target phrase $p$ and the set of text elements $O_s$ extracted by OCR for that slide. Each element $o \in O_s$ possesses recognized textual content and associated bounding polygon coordinates defining its location on the slide. The location matching function, $\text{MatchLocation}(p, O_s)$, seeks the subset of OCR elements $O_{match} \subseteq O_s$ whose polygons represent the phrase $p$. The process is configurable based on matching \textit{granularity} $G$ and \textit{method} $M$.

\subsubsection{Matching Granularity (G)}
\label{sec:granularity}
Granularity determines the level of OCR text elements used for matching and, consequently, the visual scope of the resulting highlight. This can be seen both as a technical parameter affecting precision and robustness, and as a stylistic choice influencing the user experience, mimicking how different lecturers might emphasize content. We support two levels:
\begin{itemize}
    \item \textbf{Line Granularity ($G=\text{line}$):} Matching operates on OCR elements representing complete text lines. The system attempts to find the line element(s) whose textual content contains the highlight phrase $p$. This typically results in highlighting the entire line containing the phrase. It mirrors a lecturer gesturing towards a whole line or bullet point.
    \item \textbf{Word Granularity ($G=\text{word}$):} Matching operates on OCR elements representing individual words. The system seeks a contiguous sequence of word elements whose concatenated text corresponds to the highlight phrase $p$. This allows for tighter highlights around the exact phrase, and mimics a lecturer precisely pointing to or underlining specific words.
\end{itemize}

\subsubsection{Matching Method (M)}
\label{sec:matching_method}
Given the candidate OCR elements (at the chosen granularity $G$), the matching method $M$ defines the algorithm used to identify the best match for the highlight phrase $p$. We implement several methods offering different trade-offs:

\begin{itemize}
    \item \textbf{Simple ($M=\text{simple}$):} Uses exact substring matching. For $G=\text{word}$, it identifies bounding boxes containing $p$ verbatim. This is fast and simple but inflexible to variations.

    \item \textbf{Fuzzy ($M=\text{fuzzy}$):} Employs approximate string matching, using Levenshtein distance to calculate a similarity score $\text{sim}(p, \text{candidate\_text})$. Elements with a score exceeding a threshold $\tau$ (e.g., $\text{sim} > 0.8$) are considered matches. This adds robustness to minor OCR errors or slight phrasing differences at moderate computational cost. 
\end{itemize}

While these methods handle many cases effectively, especially on text-heavy slides, they fundamentally rely on surface-level textual similarity. They struggle significantly when the highlighted phrase $p$ from the transcript relates \textit{semantically} but not \textit{literally} to the text visually present on the slide. Addressing these requires a matching method capable of deeper semantic understanding. Consider these common scenarios where simpler methods fail, and how a semantic approach can succeed:

\begin{itemize}
    \item \textbf{Abbreviations vs. Expansions:} The transcript might say "... `highlight(with respect to x)`...", while the slide visually contains the abbreviation "w.r.t. x". Simple/fuzzy matching would likely fail. A semantic approach could recognize the equivalence between the abbreviation and its full expansion.
    \item \textbf{Spoken Formula vs. Visual Notation:} The transcript verbally reads out or describes a formula (e.g., "`highlight(one minus the sum from j equals one to p of phi sub j L to the j)`") while the slide displays the corresponding compact mathematical notation (e.g., $1 - \sum_{j=1}^{p} \phi_j L^j$). Literal matching is impossible. A method capable of understanding mathematical language could link the descriptive narration to the symbolic representation.
    \item \textbf{Concept Name vs. Formula:} A slide might show an equation like $\mathcal{L}(\theta) = -\sum_{i=1}^{N} [y_i \log(\hat{y}_i) + (1-y_i) \log(1-\hat{y}_i)]$, while the transcript refers to it conceptually as the "`highlight(cross-entropy loss)`". Literal methods cannot bridge this conceptual gap. Semantic understanding could associate the common name of the concept with its mathematical formula.
    \item \textbf{OCR Misinterpretation Recovery:} OCR might misread a visually similar symbol (e.g., $\Sigma$ as 'E'). A highlight target $p$ like "`summation $\Sigma$`" would fail simple/fuzzy matching against the incorrect OCR text "E". A semantic method, potentially using surrounding context or knowledge of common errors, could infer the intended match despite the textual discrepancy.
    \item \textbf{Multiple Occurrences Disambiguation:} The term "`highlight(generator)`" appears twice on the slide. Simple/fuzzy methods might match both or only the first one found, regardless of context. By analyzing the surrounding context, a semantic method could identify which specific instance of the repeated term is being discussed.
\end{itemize}

\subsection*{LLM-based Matching ($M=\text{llm}$)}
To address these complex matching scenarios, we employ an LLM-based approach. The LLM is provided with not only the target phrase $p$ and the candidate OCR elements, but also with the surrounding context from the transcript. This allows the model to leverage semantic understanding and contextual clues for disambiguation. The core instruction given to the LLM follows this structure, presented here for clarity with an example:

\vspace{1ex}
\hrulefill
\begin{quote}
\begin{small}
\textbf{Text preceding highlight phrase:}
\textit{... Now, let's look at the general formula for ...}

\textbf{Target Highlight Phrase:}
`{Cross-entropy loss}`

\textbf{Text succeeding highlight phrase:}
\textit{... As you can see at the top, the loss 'l' is a function of the input  ...}

\vspace{1ex}
\hrulefill
\vspace{1ex}

\textbf{Candidate OCR Text Elements from Slide:}
\begin{enumerate} 
    \item Objective Function
    \item $L(\theta) = -\sum_{i=1}^{N} [y_i \log(\hat{y}\_i) + (1-y\_i)\log(1-\hat{y}_i)]$
    \item `$y_i$ is the true label`
    \item Cross-entropy
    \item[\dots]
    \item[(N)] {`text of element N`}
\end{enumerate}
\vspace{1ex}
\hrulefill
\vspace{1ex}

\textbf{Task:}
Considering the target phrase, its surrounding context, and the candidate text elements (which represent content visually present on the slide), identify the index(es) corresponding to the candidate element(s) that best match the target highlight phrase in its given context.
\end{small}
\end{quote}
\hrulefill
\vspace{1ex}

The LLM processes this combined information – the target phrase, the context clarifying its specific usage, and the list of visually present text candidates – to return a list of indices $I$. The corresponding OCR elements $O_{match} = \{ o_i \in O_s \mid i \in I \}$ are then selected.

\subsection{Timing Calculation}
\label{sec:timing_lookup}

After identifying the visual location $O_{match}$ for a highlighted phrase $p$, we determine the precise time interval $[start\_ms, end\_ms]$ for its display, ensuring synchronization with the spoken narration. This relies on the detailed timing information $T_s$ obtained from the timestamp-providing TTS module (Section \ref{sec:tts}). $T_s$ is a sequence of tuples mapping each spoken word $w_i$ to its start and end timestamps, $t_{i,start}$ and $t_{i,end}$. The timing lookup function, $\text{LookupTime}(p, T_s, \text{index})$, operates as follows:
\begin{enumerate}
    \item Identify the sequence of words $W_p = (w_{p,1}, \dots, w_{p,k})$ constituting phrase $p$ in the original transcript.
    \item Locate the index-th occurrence of the exactly corresponding contiguous subsequence $(w_i, \dots, w_j)$ within the TTS timestamp data $T_s$, where $j=i+k-1$.
    \item Extract the interval boundaries: $start\_ms = t_{i,start}$ and $end\_ms = t_{j,end}$.
\end{enumerate}
This process yields the precise interval $[start\_ms, end\_ms]$ for rendering the specific occurrence of the highlight synchronized with the generated audio.

\section{Evaluation}
\label{sec:evaluation}

\subsection{Experimental Setup}
\label{sec:eval_setup}
\textbf{Dataset:} Our evaluation utilizes a diverse dataset comprising 5 distinct university courses from different domains: Probability Theory, Financial Mathematics, Financial Technologies, Comparative Politics and Urban Energy Systems. The dataset includes a total of 100 lecture PDFs, containing approximately 5000 slides, all sourced from MIT Open CourseWare \cite{mitocw_general}. We categorize the courses to form two subsets for analysis: a \textbf{Math-Heavy} subset (Probability Theory, Financial Mathematics) characterized by frequent equations and mathematical notation, and a \textbf{Text-Heavy} subset (the remaining courses) predominantly featuring text, lists, and some diagrams/graphs. We manually annotated 1000 highlight instances across both subsets. For each sampled highlight phrase $p$, annotators inspected the OCR output $O_s$ and identified the set of word element indices $O_{true} \subseteq O_s$ constituting the correct visual representation of $p$. 

Table \ref{tab:pipeline_comparison} outlines the specific configuration used for our primary evaluation pipeline alongside corresponding open-source software (OSS) alternatives.

\begin{table}[ht]
  \centering
  \caption{Core Module Component Options.}
  \label{tab:pipeline_comparison}
  \begin{tabular}{lll}
    \toprule
    \textbf{Module} & \textbf{Model} & \textbf{Alt. OSS model} \\
    \midrule
    Narration & Gemini 2.5 Pro \cite{google_gemini25} & Llama 3.2 \cite{grattafiori2024llama3herdmodels} \\
    Alignment & Gemini 2.5 Pro \cite{google_gemini25} & Llama 3.2 \cite{grattafiori2024llama3herdmodels} \\
    OCR & Azure OCR \cite{azureocr} & Tesseract OCR \cite{tesseract} \\
    TTS & Lemonfox TTS \cite{lemonfox} & WhisperX \cite{bain2022whisperx} \\
    \bottomrule
  \end{tabular}
\end{table}

Our evaluation systematically explores the impact of two key configuration dimensions on highlight alignment accuracy:
\begin{itemize}
    \item \textbf{Granularity ($G$):} We compare performance at both `word` level and `line` level.
    \item \textbf{Matching Method ($M$):} We evaluate three distinct methods: `simple` (exact substring/sequence), `fuzzy` (Levenshtein distance), and `llm` (semantic matching via a large language model).
\end{itemize}
The subsequent sections analyze the performance across relevant combinations of these granularity levels and matching methods.

\subsection{Highlight Location Accuracy}
\label{sec:eval_word_location_accuracy}
We evaluate the performance of different matching methods at word-level granularity ($G=\text{word}$) and at line-level granularity ($G=\text{line}$). Table \ref{tab:word_level_accuracy} presents the Match Success Rate (MSR), Precision, Recall, and F1-Scores for WS, WF, and WL, broken down by content type. MSR indicates the percentage of annotated highlight instances for which the configuration successfully identified any matching OCR elements. Table \ref{tab:line_level_accuracy} presents the corresponding accuracy metrics for the line-level configurations ($G=\text{line}$), evaluated based on selecting the correct line elements.

\begin{table*}[t]
  \centering
  \caption{Word-Level Highlight Location Accuracy by Matching Method and Content Type.}
  \label{tab:word_level_accuracy}
  \begin{tabular}{lcccccccccccc}
    \toprule
    & \multicolumn{4}{c}{\textbf{Overall (N=1000)}} & \multicolumn{4}{c}{\textbf{Text-Heavy Subset}} & \multicolumn{4}{c}{\textbf{Math-Heavy Subset}} \\
    \cmidrule(lr){2-5} \cmidrule(lr){6-9} \cmidrule(lr){10-13}
    \textbf{Configuration} & \textbf{MSR(\%)} & \textbf{Prec.} & \textbf{Rec.} & \textbf{F1} & \textbf{MSR(\%)} & \textbf{Prec.} & \textbf{Rec.} & \textbf{F1} & \textbf{MSR(\%)} & \textbf{Prec.} & \textbf{Rec.} & \textbf{F1} \\
    \midrule
    WS (Word, Simple) & 62.7 & 86.0 & 53.8 & 66.2 & 85.7 & 94.1 & 68.6 & 79.3 & 41.9 & 78.6 & 43.6 & 56.1 \\
    WF (Word, Fuzzy)  & 84.7 & 52.0 & 15.2 & 23.5 & \textbf{100.0} & 67.9 & 27.1 & 38.8 & 71.0 & 31.8 &  6.9 & 11.4 \\
    WL (Word, LLM)    & \textbf{96.6} & \textbf{95.1} & \textbf{90.1} & \textbf{92.5} & 96.4 & \textbf{96.9} & \textbf{88.6} & \textbf{92.5} & \textbf{96.8} & \textbf{93.9} & \textbf{91.1} & \textbf{92.5} \\
    \bottomrule
  \end{tabular}
\end{table*}

\begin{table*}[t]
  \centering
  \caption{Line-Level Highlight Location Accuracy by Matching Method and Content Type.}
  \label{tab:line_level_accuracy}
    \begin{tabular}{lcccccccccccc}
    \toprule
    & \multicolumn{4}{c}{\textbf{Overall (N=1000)}} & \multicolumn{4}{c}{\textbf{Text-Heavy Subset}} & \multicolumn{4}{c}{\textbf{Math-Heavy Subset}} \\ 
    \cmidrule(lr){2-5} \cmidrule(lr){6-9} \cmidrule(lr){10-13}
    \textbf{Configuration} & \textbf{MSR(\%)} & \textbf{Prec.} & \textbf{Rec.} & \textbf{F1} & \textbf{MSR(\%)} & \textbf{Prec.} & \textbf{Rec.} & \textbf{F1} & \textbf{MSR(\%)} & \textbf{Prec.} & \textbf{Rec.} & \textbf{F1} \\
    \midrule
    LS (Line, Simple) & 64.4 & 73.7 & 41.8 & 53.3 & 85.7 & 83.3 & 60.6 & 70.2 & 45.2 & 57.1 & 23.5 & 33.3 \\
    LF (Line, Fuzzy)  & 83.1 & 67.3 & 49.3 & 56.9 & \textbf{100.0} & 78.6 & 66.7 & 72.1 & 67.7 & 52.4 & 32.4 & 40.0 \\
    LL (Line, LLM)    & \textbf{100.0} & \textbf{74.1} & \textbf{94.0} & \textbf{82.9} & \textbf{100.0} & \textbf{91.7} & \textbf{100.0} & \textbf{95.7} & \textbf{100.0} & \textbf{61.2} & \textbf{88.2} & \textbf{72.3} \\
    \bottomrule
  \end{tabular}
\end{table*}

\subsection{Superior Performance of LLM-based Alignment}
\label{sec:eval_analysis}

The evaluation results (Tables \ref{tab:word_level_accuracy} \& \ref{tab:line_level_accuracy}) clearly favor the LLM-based alignment methods (WL and LL). Across both word and line granularities, these semantic approaches significantly outperform methods relying on surface-level textual similarity (Simple and Fuzzy). This performance gap is particularly pronounced on the Math-Heavy subset, confirming the expectation that scenarios requiring semantic understanding (as discussed in Section \ref{sec:matching_method}) frequently arise in technical content and are poorly handled by literal matching techniques.

At the word level (Table \ref{tab:word_level_accuracy}), the LLM approach (WL) consistently achieves high accuracy (Overall F1 92.5\%), demonstrating its ability to precisely locate the intended phrase. Fuzzy matching (WF), however, proves largely unsuitable for matching sequences of OCR words (Overall F1 23.5\%). Simple matching (WS) is only viable when an exact textual match exists, leading to significantly lower recall, especially on math-heavy slides where non-literal references are common (Math-Heavy Recall 43.6\% vs. Text-Heavy Recall 68.6\%).

For line-level highlighting (Table \ref{tab:line_level_accuracy}), the LLM method (LL) again delivers the best overall performance, reliably identifying the correct line context (Overall Recall 94.0\%). Its moderate precision (Overall Prec. 74.1\%), particularly on math-heavy slides, does indicate a tendency to sometimes include adjacent or tangentially related lines. Notably, Fuzzy matching (LF) performs considerably better at the line level (Overall F1 56.9\%) than at the word level. While still significantly outperformed by the LLM, this suggests fuzzy substring matching within a line offers a usable, though limited, non-semantic baseline for broader highlighting, unlike its word-level counterpart. Simple matching (LS) again shows clear limitations, particularly on the math-heavy subset.

\subsection{Cost Analysis}
\label{sec:cost_analysis}

While the accuracy evaluation demonstrates the effectiveness of the LLM-based alignment configuration, practical adoption also depends on the economic viability of the system. This section analyzes the operational costs associated with generating video lectures using AutoLectures. We analyze the cost breakdown per component for the LLM-enabled pipeline, project the total cost for typical lecture lengths, and compare this conceptually to the estimated cost of manual production. For this analysis, we use representative public pricing available as of May 2025. The specific services and their unit costs used in this estimation are summarized in Table \ref{tab:unit_prices}.

\begin{table*}[h]
  \centering
  \caption{Unit Prices for API Services (May 2025).}
  \label{tab:unit_prices}
  \begin{tabular}{lllr}
    \toprule
    \textbf{Module} & \textbf{Provider} & \textbf{Unit} & \textbf{Price (USD)} \\
    \midrule
    OCR            & Azure Read v4   & 1000 pages & \$1.50 \\
    TTS            & Lemonfox TTS & 1M characters & \$2.50 \\
    LLM            & Gemini 2.5 Pro   & 1M input tokens  & \$1.25 \\
                   &                 & 1M output tokens & \$10.00 \\
    \bottomrule
  \end{tabular}
\end{table*}

\subsubsection{Estimated Per-Slide Cost Breakdown}
\label{sec:cost_per_slide}

\begin{figure}[ht] 
\centering
\includegraphics[width=\linewidth]{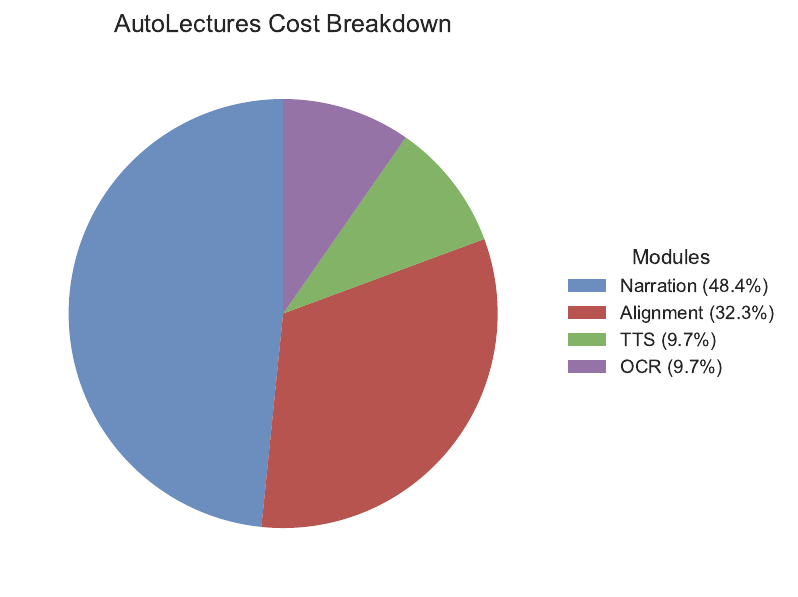}
\caption{AutoLectures cost breakdown by module (right) for a typical 60-minute lecture.}
\label{fig:cost_comparison_breakdown}
\Description{Pie chart titled 'AutoLectures Cost Breakdown' with four slices (blue largest, then red, purple, green smallest) and an external legend listing modules (TTS, Alignment LLM, Narration LLM, OCR) with percentages.}
\end{figure}

To understand the contribution of each module to the overall cost, we analyzed the average usage of each module per slide across our generated lecture dataset. Table \ref{tab:usage_and_cost_breakdown} details these measured usage averages and the resulting estimated average API cost per slide for each component, calculated using the prices in Table \ref{tab:unit_prices}. This breakdown assumes the use of the high-accuracy LLM-based alignment strategy. Figure \ref{fig:cost_comparison_breakdown} (right side) visually illustrates the relative cost contribution of each module.

\begin{table*}[ht]
  \centering
  \caption{Estimated Cost per Slide by Component. Costs calculated using prices from Table \ref{tab:unit_prices}. Usage figures are averages from generated lectures}
  \label{tab:usage_and_cost_breakdown}
  \begin{tabular}{llr} 
    \toprule
    \textbf{Module} & \textbf{Average Usage per Slide} & \textbf{Avg. Cost (USD)} \\ 
    \midrule
    OCR            & 1.0 page             & \$0.0015 \\
    TTS            & 600 characters       & \$0.0015 \\
    \midrule
    \multirow{2}{*}{Narration (LLM)} 
                   & 2000 input tokens    & \multirow{2}{*}{\$0.0075} \\ 
                   & 500 output tokens    &                           \\
    \midrule
    \multirow{3}{*}{Alignment (LLM)} 
                   & 5.0 highlights       & \multirow{3}{*}{\$0.005} \\ 
                   & 400 input tokens / highlight &                           \\
                   & 50 output tokens / highlight &                           \\
    \midrule
    \multicolumn{2}{r}{\textbf{Total}} & \textbf{\$0.0155} \\
    \bottomrule
  \end{tabular}
  \vspace{1em}
  \caption*{Costs calculated using prices from Table \ref{tab:unit_prices}. Usage figures are averages from generated lectures.}
\end{table*}

\subsubsection{Scaling to Full Lectures and Comparison to Manual Effort}
\label{sec:cost_scaling}

Extrapolating the average per-slide cost of \$0.0155 (Table \ref{tab:usage_and_cost_breakdown}) allows us to project the total cost for generating full lectures. Manually creating a narrated 60-minute video lecture of reasonable quality from existing slides often demands significant preparation and production time. A conservative estimate might place this effort in the range of \textbf{2 to 4 hours}, encompassing planning, recording, basic editing, and rendering. Assigning a value to this time (e.g., \$50-\$100 per hour for an educator or specialist) suggests a manual production cost between \textbf{\$100 and \$400}. In stark contrast, AutoLectures offers a substantial potential reduction in direct cost. Based on current pricing, the expenses for generating a comparable hour-long (~60 slide) lecture are estimated at just \textbf{\$0.93}. This cost scales approximately linearly with lecture length, meaning a shorter (~30 slide) lecture would cost around \textbf{\$0.47}, while a longer (~100 slide) one would be about \textbf{\$1.55}. As illustrated conceptually in Figure \ref{fig:cost_comparison_breakdown}, this automated approach dramatically lowers the cost barrier. While automated lecture generation is not a replacement for real lectures yet, the savings in terms of production time and associated cost appear significant, differing by roughly two orders of magnitude.

\section{Discussion}
\label{sec:discussion}

Our evaluation demonstrates that AutoLectures can effectively automate the generation of narrated video lectures with synchronized visual highlights. The results confirm that the LLM-based alignment strategy achieves high location accuracy, significantly outperforming simpler methods, particularly on slides with complex mathematical notation or requiring semantic interpretation (Section \ref{sec:eval_analysis}). Furthermore, our cost analysis reveals that generating lectures using this high-accuracy pipeline is remarkably economical, with estimated costs under \$1 for a typical hour-long lecture (Section \ref{sec:cost_analysis}).

Beyond automation, a key goal of AutoLectures is to enhance the pedagogical value of generated videos. By incorporating dynamically synchronized highlights, the system directly operationalizes the Signalling Principle from multimedia learning theory~\cite{signallingprinciple}. The demonstrated accuracy of the LLM-based alignment ensures that these automatically generated cues can effectively guide viewer attention to relevant textual information, even in challenging cases like matching concept names to formulas or for disambiguating terms using surrounding context.

The significant cost reduction compared to manual production (roughly two orders of magnitude) has the possibility to have profound practical implications. It dramatically lowers the barrier for educators to create video resources from their existing slide materials. This can free up valuable time often spent on the repetitive task of recording and editing standard lectures, allowing educators, researchers, and other experts to dedicate more focus to primary activities such as research, curriculum development, and direct student interaction. The low cost and automated nature also enable the scalable production and updating of video lectures, potentially increasing the accessibility and reach of educational content across diverse settings. While not a replacement for live teaching, AutoLectures offers a practical tool to supplement traditional methods and for enhancing asynchronous learning opportunities.

\begin{table*}[ht]
  \centering
  \caption{Appendix: Word-Level Highlight Location Accuracy Comparison Across Different LLMs ($G=\text{word}, M=\text{llm}$). Citations refer to model documentation or announcements in the cases where no papers are available.}
  \label{tab:appendix_llm_accuracy}
  \begin{tabular}{l cccc cccc cccc}
    \toprule
     & \multicolumn{4}{c}{\textbf{Overall}} & \multicolumn{4}{c}{\textbf{Text-Heavy Subset}} & \multicolumn{4}{c}{\textbf{Math-Heavy Subset}} \\
    \cmidrule(lr){2-5} \cmidrule(lr){6-9} \cmidrule(lr){10-13}
    \textbf{Alignment LLM} & \textbf{MSR(\%)} & \textbf{Prec.} & \textbf{Rec.} & \textbf{F1} & \textbf{MSR(\%)} & \textbf{Prec.} & \textbf{Rec.} & \textbf{F1} & \textbf{MSR(\%)} & \textbf{Prec.} & \textbf{Rec.} & \textbf{F1} \\
    \midrule
    GPT-4.1~\cite{openai_gpt41}              & \textbf{100.0} & 81.8 & 90.3 & 85.9          & \textbf{100.0} & 81.8 & 93.3 & 87.2          & \textbf{100.0} & 81.9 & 87.9 & 84.8 \\
    OpenAI o3~\cite{openai_o3}                & 98.0           & 85.6 & 90.3 & 87.9          & \textbf{100.0} & 88.5 & \textbf{96.7} & 92.4          & 95.8           & 83.0 & 85.2 & 84.1 \\
    Gemini 2.5 Pro~\cite{google_gemini25}  & 94.0           & \textbf{96.0} & 90.3 & \textbf{93.1} & 94.2           & \textbf{98.2} & 89.2 & \textbf{93.4} & 93.8           & \textbf{94.4} & \textbf{91.3} & \textbf{92.8} \\
    Gemini 2.5 Flash~\cite{google_gemini25}& 94.0           & 93.3 & 87.7 & 90.4          & 94.2           & 97.2 & 87.5 & 92.1          & 93.8           & 90.3 & 87.9 & 89.1 \\
    DeepSeek V3~\cite{deepseekai2025deepseekv3technicalreport}      & 99.0           & 82.2 & \textbf{90.7} & 86.2 & \textbf{100.0} & 80.3 & 91.7 & 85.6          & 97.9           & 83.8 & 89.9 & 86.7 \\
    Grok 3~\cite{xai_grok3}                & 96.0           & 81.7 & 86.2 & 83.9          & \textbf{100.0} & 83.9 & 95.8 & 89.5          & 91.7           & 79.6 & 78.5 & 79.1 \\
    \bottomrule
  \end{tabular}
\end{table*}

\subsection{Limitations}

While AutoLectures demonstrates a viable approach to automated video lecture generation with highlights, several limitations should be acknowledged.

The core highlight alignment mechanism, while effective for text, primarily relies on matching narrated phrases to text elements identified by OCR. Consequently, its ability to handle references to non-textual visual content is currently limited. For example, aligning narration like "...focus on the upper-left quadrant of the graph..." or "...this specific neuron cluster..." to the correct visual region without corresponding text labels poses a significant challenge. Furthermore, even with LLM-based methods, highlight location accuracy is not perfect, meaning occasional errors in placement or missed highlights can occur.

Another limitation is that the system currently implements only one form of visual guidance: rectangular bounding box highlights around existing text. It does not generate other potentially beneficial cue types. For example, it cannot draw arrows to point to specific elements, nor can it replicate the dynamic free-form annotations (e.g writing brief notes directly on the slide) that human presenters often use to elaborate on or connect ideas visually. These alternative forms of visual interaction could be more effective for certain types of content or pedagogical goals, such as indicating relationships between elements or illustrating a process step-by-step.

Finally, and most significantly from a pedagogical perspective, this work focused on the technical feasibility, cost, and specifically the location accuracy of the highlight alignment module. We did not conduct user studies to empirically evaluate the actual impact of the automatically generated videos on learner outcomes (e.g., retention or transfer performance). While the design is motivated by the Signaling Principle, further research involving human learners is required to confirm whether the generated highlights, along with the AI narration quality and highlight token choices, translate into measurable learning benefits compared to unguided videos or manually created content.

\subsection{Future Work}

Based on the limitations identified, several promising avenues for future work emerge. One interesting direction involves exploring an alternative architecture that utilizes multimodal LLMs even more to potentially unify transcript generation and visual grounding. Instead of the current multi-stage pipeline where a transcript with highlight tokens is generated first, followed by separate OCR and alignment steps, the multimodal LLM could potentially process the input slides and generate the transcript while simultaneously outputting the geometric coordinates or parameters for desired visual cues, directly associated with the relevant generated phrases. For example, when generating the phrase "cross-entropy loss", the model would concurrently determine the location of the corresponding formula or text on the input image and output its bounding box coordinates. This paradigm could inherently address current limitations in handling non-textual references (as the model could directly ground phrases like "the upper-left quadrant" to image coordinates) and enable the generation of diverse cue types (outputting parameters for arrows or simple annotations instead of just boxes). Such an end-to-end approach would bypass the need for explicit highlight tokens and the subsequent alignment module. Realizing this vision depends on future advancements in the fine-grained visual grounding and instruction-following capabilities of multimodal models, but it represents a potentially significant simplification and enhancement. 

Another area for future work is the empirical evaluation of pedagogical effectiveness. While this paper established the technical feasibility, alignment accuracy, and cost-efficiency of AutoLectures, it did not measure the actual impact of the generated videos on human learning. Controlled experiments are needed to rigorously assess whether the automatically generated synchronized highlights, motivated by the Signaling Principle, demonstrably improve learning outcomes. Such studies should compare learner performance (measured via standard retention and transfer tests) between groups viewing AutoLectures videos \textit{with} highlights versus identical videos generated \textit{without} highlights. Further comparisons against manually produced video lectures covering the same slide content could also provide valuable benchmarks, although controlling for confounding factors like presenter style and specific cue choices presents methodological challenges.


\appendix

\section{Comparison of LLMs for Word-Level Alignment}
\label{sec:appendix_llm_comparison}

Given that the LLM-based method ($M=\text{llm}$) demonstrated superior performance for word-level highlight alignment (Configuration WL, Section \ref{sec:eval_analysis}), we conducted a supplementary analysis to investigate the impact of the specific Large Language Model choice within this configuration ($G=\text{word}, M=\text{llm}$). This comparison helps assess whether the high accuracy observed generalizes across other contemporary models beyond the primary LLM used in our main evaluation (Gemini 2.5 Pro).

We evaluated several distinct LLMs available as of May 2025 on the word-level alignment task using the AutoLectures-1K dataset subset (N=100). Table \ref{tab:appendix_llm_accuracy} reports the standard location accuracy metrics (MSR, Precision, Recall, F1-Score) across the Overall, Text-Heavy, and Math-Heavy subsets for each tested model.

\noindent 
The results in Table \ref{tab:appendix_llm_accuracy} indicate that strong performance on the word-level alignment task is achievable across multiple contemporary LLMs. While all tested models demonstrate relatively high accuracy, Gemini 2.5 Pro stands out, achieving the highest F1-score overall and on both subsets, driven primarily by superior precision, particularly on the challenging Math-Heavy content. Gemini 2.5 Flash also performs competitively, especially considering its efficiency advantages. OpenAI o3 and DeepSeek V3 show strong results as well, with OpenAI o3 achieving the highest recall on the Text-Heavy subset and DeepSeek V3 achieving the highest overall recall and strong F1 on Math-Heavy content. GPT-4.1 and Grok 3 deliver solid performance but lag slightly behind the top performers on this specific task based on F1-score. This comparison provides confidence in the general effectiveness of using LLMs for semantic highlight alignment and further supports the use of Gemini 2.5 Pro in our main evaluation. Note that this analysis focuses solely on alignment accuracy; relative cost and inference latency were not evaluated here.


\bibliographystyle{ACM-Reference-Format}
\bibliography{references}

\end{document}